\documentclass[preprint,12pt]{elsarticle}

\usepackage{amssymb}
\usepackage{amsmath}

\usepackage{multirow}
\usepackage{booktabs} %
\usepackage{pifont}
\usepackage{adjustbox}
\usepackage[section]{placeins} %
\usepackage{float} %
\usepackage{tabularx} %

\journal{}
\makeatletter
\def\ps@pprintTitle{%
  \let\@oddhead\@empty   %
  \let\@evenhead\@empty  %
  \def\@oddfoot{\hfill\thepage\hfill}  %
  \def\@evenfoot{\hfill\thepage\hfill} %
}
\makeatother
\makeatother
\begin{document}

\begin{frontmatter}
\title{Deformation-aware Temporal Generation for Early Prediction of Alzheimer's Disease}
\author[HQU,LAB]{Xin Hong}
\author[HQU]{Jie Lin}
\author[HQU]{Minghui Wang}
\affiliation[HQU]{organization={the College of Computer Science and Technology,Huaqiao University},
            addressline={Jimei Road 668}, 
            city={Xiamen},
            postcode={361021}, 
            country={China}}

    \affiliation[LAB]{organization={the Key Laboratory of Computer Vision and Machine Learning in Fujian Province},
    addressline={Jimei Road 668}, 
            city={Xiamen},
            postcode={361021}, 
            country={China}
            }

\begin{abstract}
Alzheimer's disease (AD), a degenerative brain condition, can benefit from early prediction to slow its progression. As the disease progresses, patients typically undergo brain atrophy. Current prediction methods for Alzheimer's disease largely involve analyzing morphological changes in brain images through manual feature extraction. This paper proposes a novel method, the Deformation-Aware Temporal Generative Network (DATGN), to automate the learning of morphological changes in brain images about disease progression for early prediction. Given the common occurrence of missing data in the temporal sequences of MRI images, DATGN initially interpolates incomplete sequences. Subsequently, a bidirectional temporal deformation-aware module guides the network in generating future MRI images that adhere to the disease's progression, facilitating early prediction of Alzheimer's disease. DATGN was tested for the generation of temporal sequences of future MRI images using the ADNI dataset, and the experimental results are competitive in terms of PSNR and MMSE image quality metrics. Furthermore, when DATGN-generated synthetic data was integrated into the SVM vs. CNN vs. 3DCNN-based classification methods, significant improvements were achieved from 6. 21\% to 16\% in AD vs. NC classification accuracy and from 7. 34\% to 21. 25\% in AD vs. MCI vs. NC classification accuracy. The qualitative visualization results indicate that DATGN produces MRI images consistent with the brain atrophy trend in Alzheimer's disease, enabling early disease prediction. 
\end{abstract}




\end{frontmatter}




\section{Introduction}
\label{sec:introduction}

Alzheimer's disease is a progressively deteriorating neurological condition that leads to a gradual decline in cognitive function, resulting in impairments in memory, thinking, and behavior. Early prediction is crucial to enable doctors and patients to implement timely interventions and treatments to improve patient quality of life. Currently, the main methods for predicting Alzheimer's disease using structural Magnetic Resonance Imaging (MRI) include predicting future disease status (clinical scores)\cite{rallabandi2023deep, liu2014multimodal, suk2015latent, jung2021deep, kim2021interpretable, chincarini2016integrating, cui2018longitudinal, er2020predicting, wang2018predictive}and predicting future brain images\cite{lin2021bidirectional, jung2023conditional, zhou2019latent, shin2020gandalf}. Given that MRI provides more information, predicting future MRI images has greater diagnostic significance compared to predicting the disease state. 

Prediction-based methods for generating Alzheimer's disease MRI images are primarily categorized into non-temporal brain image generation methods\cite{lin2021bidirectional,jung2023conditional,zhou2019latent} and temporal brain image generation methods\cite{fan2022tr,ning2020ldgan}. Non-temporal MRI image generation methods typically employ image-to-image generation techniques to produce individual brain images, e.g., incorporating AD class label information, patient's clinical data, inter-modality correlations, and other features. However, these methods do not consider the correlation between time series and brain atrophy in time-series MRI images, making it challenging to model diseases progression. 

In contrast, temporal brain image generation methods generate brain images by exploiting the temporal correlation of time-series data, utilizing techniques such as recurrent neural networks\cite{fan2022tr} and multiple generators\cite{ning2020ldgan}, etc.. Despite the higher costs associated with data collection and processing, temporal data offers more comprehensive insights\cite{bhagwat2018modeling}, including disease progression and changes, thereby enhancing disease prediction performance.   

The generation of temporal brain images can depict the potential evolution of Alzheimer's disease progression. As the disease advances, individuals develop characteristic brain atrophy, primarily evidenced by alterations in brain anatomical structures associated with Alzheimer's pathology, such as ventricles, hippocampal shape, cortical thickness, and brain volume\cite{kramer2007longitudinal}. Deformation information plays a critical role in capturing the aging trajectory of Alzheimer's disease progression, and most MRI-based prediction methods for Alzheimer's disease\cite{hong2022combined} rely on manually extracted morphological changes. However, existing temporal MRI brain image generation methods\cite{fan2022tr,ning2020ldgan} seldom explore the characteristic atrophy of MRI brain image morphology during disease progression. 

To capture dynamic changes in longitudinal brain image sequences, this paper introduces a generative model named DATGN for future temporal MRI data. This model estimates morphological changes between frames in multiple temporal sequences with missing values, learns deformation fields of characteristic brain changes brain over time and the imaging representation of MRI, thereby generating future temporal MRI images to enable early prediction of Alzheimer's disease.  

\section{Related Work}

Temporal MRI image sequences often suffer from missing data due to examination discontinuity, while video sequences usually feature image continuity. Consequently, differences exist in data feature representation and sequence continuity between time-series MRI and video sequences, hindering direct application of video prediction technologies to Alzheimer's disease prediction. Time-series image generation technologies, as a branch of time-series prediction, are divided into two categories, i.e., direct generation and mapping generation. 

\subsection{Direct Generation Methods}

Direct generation methods\cite{shi2015convolutional, byeon2018contextvp, ye2022vptr, wang2022predrnn} tend to directly predict pixels in future temporal frames without modeling the scene dynamics. They achieve sequence image generation by modeling the sequence image and learn the next frame of the sequence directly. 

Srivastava et al.\cite{byeon2018contextvp} proposed an autoencoder (AE) based method that directly predicts the next frame of a video by processing temporal inputs with a Sequence-to-Sequence approach. To effectively utilize the spatial correlations in videos, Shi et al. introduced an end-to-end model that captures spatiotemporal features using ConvLSTM\cite{shi2015convolutional}, thereby predicting future pixels. Similarly, the study\cite{wang2022predrnn} used stacked RNNs to obtain future image sequences. However, such RNN-based\cite{medsker2001recurrent} methods primarily focus on capturing changes between adjacent moments, lacking sufficient modeling capability for long-term dependencies. 
Therefore, direct image generation methods have limitations in future temporal MRI generation under data missing conditions.  

In the medical field, existing direct generation methods\cite{lin2021bidirectional, jung2023conditional, zhou2019latent, shin2020gandalf, ning2020ldgan} have been applied to predict the future changes of the human brain. For example, LDGAN\cite{ning2020ldgan} utilizes multiple generators to generate missing MRI images. However, the number of generators in LDGAN and training complexity increase linearly with the length of the sequence data. TR-GAN\cite{fan2022tr} employs recurrent neural networks (RNNs) to connect MRI data from multiple past time points and iteratively generates future MRI images. This incremental generation method causes the input brain image sequence to have different sizes with each iteration. Similarly, the study \cite{xia2021learning} leverages the multi-dimensional information of time-series MRIs to generate future MRI images.

\subsection{Mapping Generation Methods}

Mapping generation methods\cite{ kwon2019predicting, oliu2018folded, villegas2017decomposing, denton2017unsupervised, liang2017dual} not only learn visual appearance information, but also introduce scene dynamic models to constrain image generation, thereby improving the network's perception capability of temporal changes.  

In the field of video applications, some methods use a combination of dynamic temporal changes and static structural information. For example, Kwon et al.\cite{kwon2019predicting} and Oliu et al. \cite{oliu2018folded} considered bidirectional temporal flow field features. Assuming that the image sequence is symmetric in time, they used forward mapping and backward mapping to predict it, ensuring the temporal consistency of the video prediction results. Hu et al.\cite{hu2019novel} considered a training method for modeling temporal change in the flow field and proposed a temporal consistency loss for training. This loss incorporates temporal changes in the encoding space for joint iterations, in order to generate temporal results that are more consistent with the time variations.

On the other hand, some methods generate future video frames by separating the dynamic temporal changes from static structural information. MCNet\cite{villegas2017decomposing} models the scene dynamics and visual appearance separately and then fuse them in an end-to-end model. 
Denton et al.\cite{denton2017unsupervised} proposed a training strategy based on adversarial loss, which models the losses of scene dynamics and visual appearance information separately. 
To improve the model's capability to predict the variability of future videos, Liang et al.\cite{liang2017dual} integrated future frame prediction and flow field estimation into a unified architecture using a shared probabilistic motion encoder to achieve future video generation.

Existing mapping generation methods often handle video frames with equal time intervals, while in the medical imaging field, mapping generation applications for time-series MRI data with missing data have not been thoroughly explored. 
Although direct generation methods have achieved certain results in modeling the development of brain images in the progression of the disease, it is often difficult to learn semantic-level changes in direct generation methods due to the uncertainty of the future, in which low-level details of temporal changes are often learned\cite{oprea2020review}. Considering the application of deformation fields in MRI-based medical image registration tasks\cite{lv2022joint}, mapping generation methods can be considered as a feasible solution for temporal modeling of MRI images in Alzheimer's disease. 

\section{Method}
\begin{figure}[ht]
\centering
\includegraphics[width=5in]{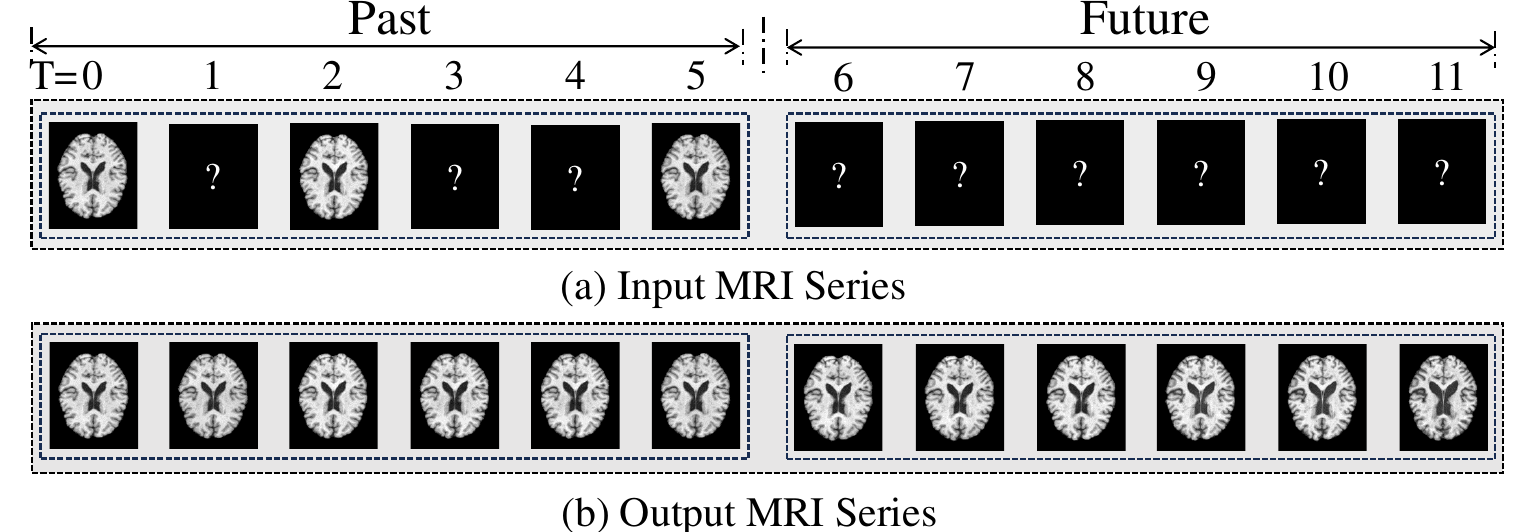}
\caption{The prediction of the progression of Alzheimer's disease involves generating future brain image sequences based on past brain image sequences. (a) represents the temporal sequence of past MRI brain images of a patient with missing data. In subfigure (b), the left part represents the temporal sequence after the missing data has been generated through an interpolation method, and the right part represents the future time-point sequence generated through a prediction method.}
\label{fig.task}
\end{figure}

In the task of generating Alzheimer's disease images, as shown in Figure \ref{fig.task}, missing data may occur due to possible interruptions or uneven examination time intervals in the examination process. 

\begin{figure}[ht]
\centering
\includegraphics[width=5in]{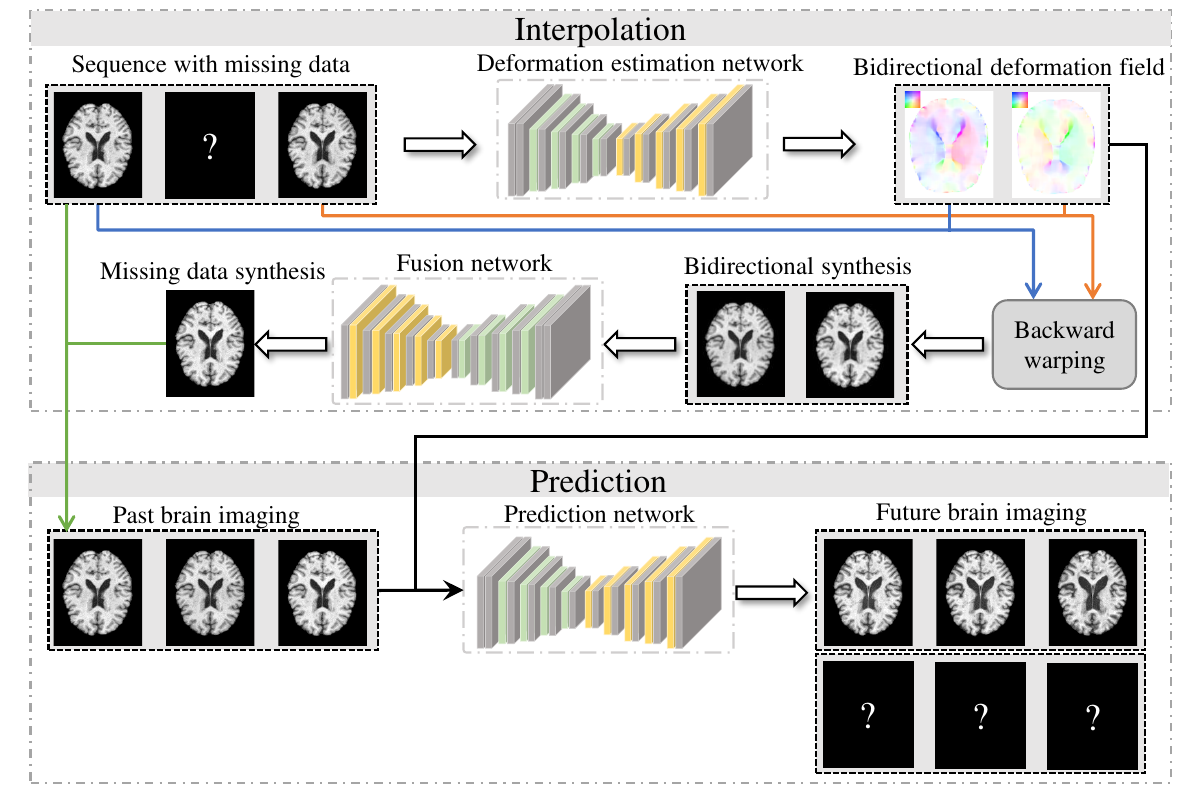}
\caption{DATGN consists of two modules: the temporal interpolation module and the temporal prediction module. The temporal interpolation module contains a deformation field estimation network and a interpolation network for completing the missing data; The temporal prediction module contains a prediction network for generating the temporal sequence of future brain images.}
\label{fig.method_overview}
\end{figure}

In order to discover the development trend of Alzheimer's disease and achieve early prediction, this paper proposes the generation of future brain image sequences using DATGN. In this approach, the deformation field is used as the core module for modeling temporal data, which can capture the morphological changes of brain images in spatial dimension, making the synthesized brain images more consistent with the temporal changes of brain lesions. Figure \ref{fig.method_overview} illustrates the algorithmic flow of DATGN, which consists of the temporal brain image interpolation module and the temporal brain image generation module. 

\subsection{The Temporal Interpolation Module}

The existence of different time gaps in the series of brain images can hinder the effectiveness of temporal modeling. Consequently, employing the deformation field mapping interpolation technique to fill in the missing data and create a complete temporal sequence may enhance the modeling of dynamic brain activity changes. This approach utilizes a deformation field estimation network alongside an interpolation network to assess alterations in brain images and forecast missing data based on these changes. 

Optical flow interpolation is a commonly used temporal interpolation method, which can insert missing frames into a temporal sequence by mapping generation, thus constructing a high frame rate temporal sequence. Existing learning-based methods\cite{jiang2018super, kalluri2023flavr, dosovitskiy2015flownet} are trained by supervising the flow field in such a way that the model can estimate the change in the flow field between two frames. However, it is difficult to apply supervised learning to temporal sequences of brain images because of the labor-intensive task of labeling temporal sequences of brain images with dense deformation field labels. However, there are discrete missing data in temporal sequences of brain images, and the distribution of frame-to-frame flow fields in temporal sequences of brain images is uneven compared to continuous temporal sequences of videos. Time-series modeling has been extensively studied in equidistantly sampled time series, but less in such irregularly sampled time series, and the modeling methods for such time series differ significantly from those for equidistantly sampled ones. 

To address the issues of absent flow field labels and unevenly missing data in time sequences, this paper presents a self-supervised training algorithm DATGN. This algorithm estimates variations in the temporal sequence through deformation fields and performs interpolation in the time series of brain images utilizing the detected change features. DATGN uses a bidirectional deformation field interpolation method to complete missing data in temporal sequences of brain images. 

Considering the characteristics of missing data in a temporal sequence of brain images, the available time point data in the sequence are used to infer the temporal information of preceding and subsequent frames to model the temporal sequence.
The DATGN temporal interpolation module comprises a deformation field estimation network, an interpolation network, and their corresponding loss functions. As shown in Figure 3, the deformation field estimation network is designed to estimate the bidirectional deformation fields ${I_{t{\rightarrow }t-1}^2}$ and ${I_{t{\rightarrow }t+1}^2}$, while the interpolation network combines the mapping frames from both directions to produce the final intermediate frame ${I_t^{pred}}$. The loss functions are used to train the deformation field network and the interpolation network. 

\begin{figure}[htp]
\centering
\includegraphics[width=5in]{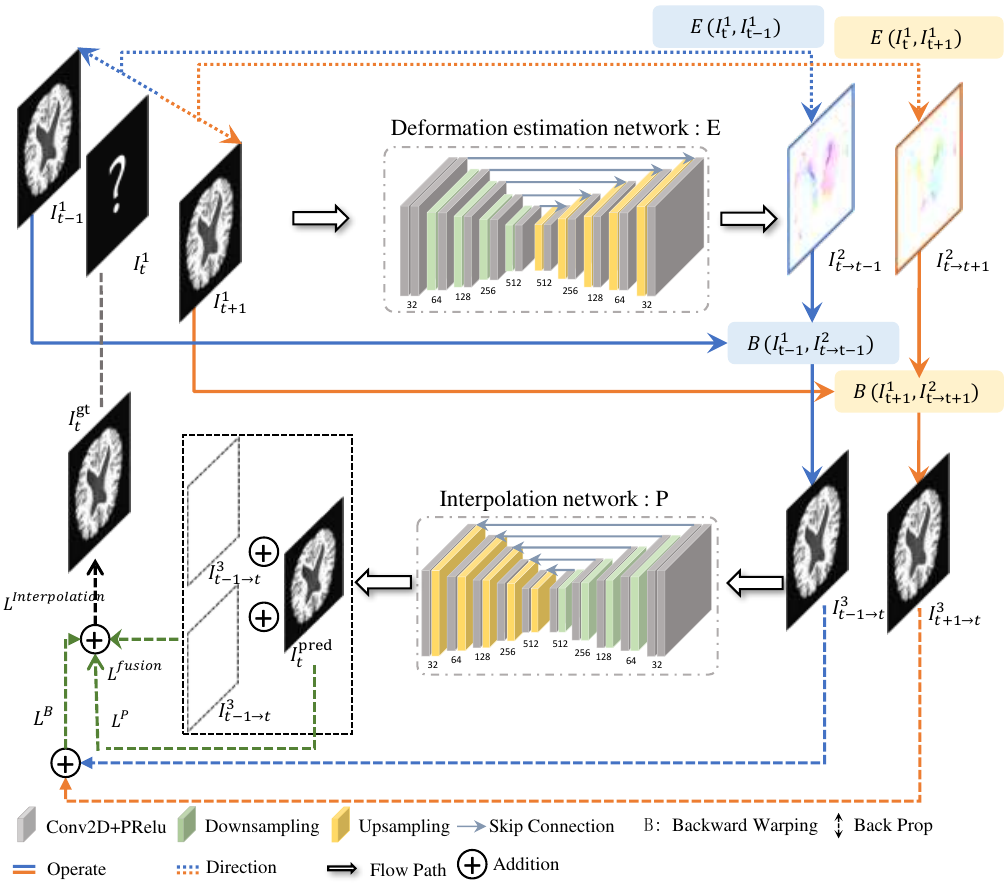}
\caption{The temporal interpolation module contains a deformation estimation network E, a interpolation network P and loss functions, wherein the deformation field estimation network $E$ estimate the inter-frame deformation fields ${I_{t{\rightarrow }t-1}^2}$,${I_{t{\rightarrow }t+1}^2}$ from the temporal brain images inputs ${I_{t-1}^1}$, ${I_{t+1}^1}$ ; and backward warping function B combines the original images ${I_{t-1}^1}$,${I_{t+1}^1}$ and the inter-frame deformation fields  ${I_{t{\rightarrow }t-1}^2}$,${I_{t{\rightarrow }t+1}^2}$ to estimated bi-direction images $ I_{t-1 \rightarrow t}^{3}$ ,$I_{t+1 \rightarrow t}^{3}$;
the interpolation network P combines $ I_{t-1 \rightarrow t}^{3}$,$I_{t+1 \rightarrow t}^{3}$ to generate the predicted intermediate frame $I_{t-1}^{pred}$, $I_{t+1}^{pred}$;${L^{Interpolation}}$ is global loss function,combine by $L^{B}$,$L^{P}$ and $L^{fusion}$, $L^{B}$ minimizing the difference between the true frame ${{I}_{t}^{gt}}$ with $ I_{t-1 \rightarrow t}^{3}$,$I_{t+1 \rightarrow t}^{3}$,$L^{P}$ minimizing the difference between the true frame ${{I}_{t}^{gt}}$ with $I_{t}^{pred}$,$L^{fusion}$ minimizing the difference between the true frame ${{I}_{t}^{gt}}$ with the combination of $ I_{t-1 \rightarrow t}^{3}$ and $I_{t}^{pred}$,$I_{t+1 \rightarrow t}^{3}$ and $I_{t}^{pred}$.
}
\label{fig.interpolation}
\end{figure}

 The DATGN model, as shown in Figure \ref{fig.interpolation}, given two frames of input images ${I_{t-1}^{1}}$ and ${I_{t+1}^{1}}$ in a temporal sequence, our objective is to synthesize the predicted frame ${I_{t}^{1}}$ for the missing moment ${t}$. DATGN initially estimates the bidirectional interframe deformation fields ${I_{t \rightarrow t-1}^{2}}$ and ${I_{t \rightarrow t+1}^{2}}$ for the input temporal sequence.  By applying bidirectional inverse remapping transformations to the deformation fields and the original frames, the intermediate brain image frames ${I_{t-1 \rightarrow t}^{3}}$ and ${I_{t+1 \rightarrow t}^{3}}$ from both directions are obtained. Finally, by fusing the intermediate frames of the brain images in both directions through the interpolation network, the predicted frame ${I_{t}}$ at time $t$ is generated.

\subsubsection{Deformation Estimation Network E}

The DATGN deformation field estimation network adopts an encoder-decoder architecture and utilizes skip connections to connect features between the encoder and decoder at the same spatial resolution. The encoder consists of 5 levels. Each level is composed of a convolutional layer and a PReLU layer (in grey). Except for the last level, an average pooling layer with a stride of 2 is used at the end of each level to reduce the spatial dimensions (in green). The decoder also consists of 5 levels. At the beginning of each level, a bilinear upsampling layer is used to double the spatial dimensions (in yellow), followed by a convolutional layer and a PReLU layer. Since brain imaging data typically have long-term temporal dependencies, i.e., complex correlations exist between the data at the current time point and the data at previous time points, when modeling deformations in the sequence, the model needs to use larger convolutional kernels in the local feature extraction stage of the encoder to capture long-range deformations. Therefore, the convolutional kernel size used in the convolutional layers gradually decreases from local to global. 


DATGN first estimates the deformation field through the deformation field estimation network $E$, and then outputs the deformation fields ${I_{t \rightarrow t-1}^{2}}$ and ${I_{t \rightarrow t+1}^{2}}$ in both directions from ${I_{t}^{1}}$ to ${I_{t-1}^{1}}$ and from ${I_{t}^{1}}$ to ${I_{t+1}^{1}}$, as shown in Eq. \ref{eq.estimator}.

\begin{equation}
\label{eq.estimator}
\begin{aligned}
I_{t{\rightarrow }t-1}^{2} = E(I_{t+1}^{1}, I_{t-1}^{1},{\alpha}_{t{\rightarrow }t-1})
\\
I_{t{\rightarrow }t+1}^{2} = E(I_{t-1}^{1}, I_{t+1}^{1},{\alpha}_{t{\rightarrow }t+1})
\end{aligned}
\end{equation}
where $E$ denotes the deformation field estimation network, $I_{t}^{1}$ is predict image which initialized as a zero matrix; $I_{t-1}^1$ and $I_{t+1}^1$ denote the input image frames, $I_{t{\rightarrow }t-1}^2$ and ${\alpha}_{t{\rightarrow }t-1}$ denotes the deformation field and parameter from the moment of $t$ to the moment of $t-1$; $I_{t{\rightarrow }t+1}^2$ and ${\alpha}_{t{\rightarrow }t+1}$ denotes the deformation field and parameter from the moment of $t$ to the moment of $t+1$.

\subsubsection{Backward Warping Function B}

The generated deformation fields ${I_{t \rightarrow t-1}^{2}}$ and ${I_{t \rightarrow t+1}^{2}}$ are used to perform backward warping operations ${B}$, which is bilinear interpolation, with the corresponding original frames ${I_{t-1}^1}$ and ${I_{t+1}^1}$ to obtain the bidirectional predicted frames ${I_{t-1 \rightarrow t}^{3}}$ and ${I_{t+1 \rightarrow t}^{3}}$, as shown in Eq.\ref{eq.backward}:
\begin{equation}
\label{eq.backward}
\begin{aligned}
\\
I_{t-1 \rightarrow t}^{3} = {B(I_{t-1}^1, I_{t{\rightarrow }t-1}^{2})}
\\
I_{t+1 \rightarrow t}^{3} = {B(I_{t+1}^{1}, I_{t{\rightarrow }t+1}^2)}
\end{aligned}
\end{equation}

The back warping loss $L^{B}$ acts on the deformation field operations of the bidirectional deformation field estimation network $E$ by calculating the discrepancy between the bidirectional mapping frames ${I_{t-1 \rightarrow t}^{3}}$ and ${I_{t+1 \rightarrow t}^{3}}$ and the true brain image label ${I_{t}^{gt}}$. The warping loss $L^{B}$ is calculated using the formula in Eq .\ref{eq.lwloss}:

\begin{equation}
\label{eq.lwloss}
\begin{aligned}
L^{B} = \lVert I_{t}^{gt}-I_{t-1 \rightarrow t}^{3}\rVert_1 + \lVert I_{t}^{gt}-I_{t+1 \rightarrow t}^{3}\rVert_1
\end{aligned}
\end{equation}
where ${I_{t}^{gt}}$ represents the true brain image.

\subsubsection{Interpolation Network P }

The interpolation network $P$, as the deformation field network $E$, also uses an encoder-decoder structure, and their backbone network structures are the same. P accepts the bidirectional predicted input frames ${I_{t-1 \rightarrow t}^{3}}$, ${I_{t+1 \rightarrow t}^{3}}$, and outputs the prediction frame $I_t^{pred} $ in time t, as shown in the Eq. \ref{eq.P} 

\begin{equation}
\label{eq.P}
\begin{aligned}
I_t^{pred} = {P(I_{t-1 \rightarrow t}^{3},I_{t+1 \rightarrow t}^{3})}
\end{aligned}
\end{equation}

The loss of reconstruction $L^{P}$ uses the intermediate predicted frame $I_t^{pred}$ generated by the interpolation network P at the moment t. The loss is computed by minimizing the difference between $I_t^{pred}$ and $I_{t}^{gt}$  by backpropagation. The calculation of the reconstruction loss $L^{P}$ is indicated by the formula in Eq. \ref{eq:interp2}:

\begin{equation}
\label{eq:interp2}
L^{P} = \lVert I_{t}^{gt}-I_t^{pred}\lVert_1
\end{equation}

Given that the brain structure of each subject is not exactly the same, in order to endow the model with adaptability to the apparent brain structure, the temporal interpolation module employs a linear fusion loss $L^{fusion}$ to further constrain the deformation estimation network and remapping, generating realistic brain images $I_t^{fusion}$. This is achieved using intermediate mapping frames $I_{t}^{pred}$from two directional brain images $I_{t-1 \rightarrow t}^{3}$ and $I_{t+1 \rightarrow t}^{3}$ to produce the intermediate fused frame $I_{t-1}^{fusion}$,$I_{t+1}^{fusion}$. The calculation process is shown in Eq. \ref{eq.fuse}:  

\begin{equation}
\label{eq.fuse}
\begin{aligned}
\\
I_{t-1}^{fusion} &= I_{t-1 \rightarrow t}^{3} + I_{t}^{pred}
\\
I_{t+1}^{fusion} &= I_{t+1 \rightarrow t}^{3} + I_{t}^{pred}
\\
L^{fusion} &= \lVert I_{t}^{gt}-I_{t-1}^{fusion}\rVert_1 + \lVert I_{t}^{gt}-I_{t+1}^{fusion}\rVert_1
\end{aligned}
\end{equation}
where the parameter ${{\alpha}_0}$ controls the temporal consistency between the two input images ${I_{t-1 \rightarrow t}^{3}}$ and ${I_{t+1 \rightarrow t}^{3}}$. ${\odot}$ represents element-wise matrix multiplication. ${I_{t}^{fusion} }$ represents the fused frame at moment t. 

\subsubsection{The Global Loss Function}

The temporal module of interpolation of the brain image adopts a self-supervised training method, and its global loss function ${L^{Interpolation}}$ is a linear combination of three terms, as shown in the Eq. \ref{eq:interp}:

\begin{equation}
\label{eq:interp}
L^{Interpolation} =  \lambda^{B} L^{B}+\lambda^{P} L^{P}  + \lambda^{fusion} L^{fusion}
\end{equation}
where $\lambda^{B}$ represents the linear coefficient of the warping loss, $L^{B}$ represents the warping loss; 
$\lambda^{P}$ represents the linear coefficient of the reconstruction loss of interpolation network P  $L^{P}$ represents the reconstruction loss; 
$\lambda^{fusion} $ represents the linear coefficient of the fusion loss, $L^{fusion} $ represents the fusion loss.  

Every component of $L^{Interpolation}$ is differentiable, thus DATGN can perform end-to-end self-supervised training. This allows DATGN to not only obtain high quality interpolation network weights at moment t, but also to acquire network weights that can accurately estimate deformation fields.

\subsection{The Temporal Prediction Module}

Due to the different data distributions of the static structural information of the brain images and the deformation information associated with disease progression in the time series, modeling them separately can better utilize the distinct features of these two types of information. The spatial encoder is used to capture static structural information, while the deformation field encoder is used to capture the deformation information, thereby assisting the model in generating future brain images that align with the evolving disease state. 
The DATGN temporal prediction module is based on a deformation-aware temporal generation technique to predict future brain images in a time series. It consists of a spatial encoder, a deformation field encoder, a DT-Module, a decoder, and a loss function.  

\begin{figure*}[htp]
\centering
\includegraphics[width=\textwidth]{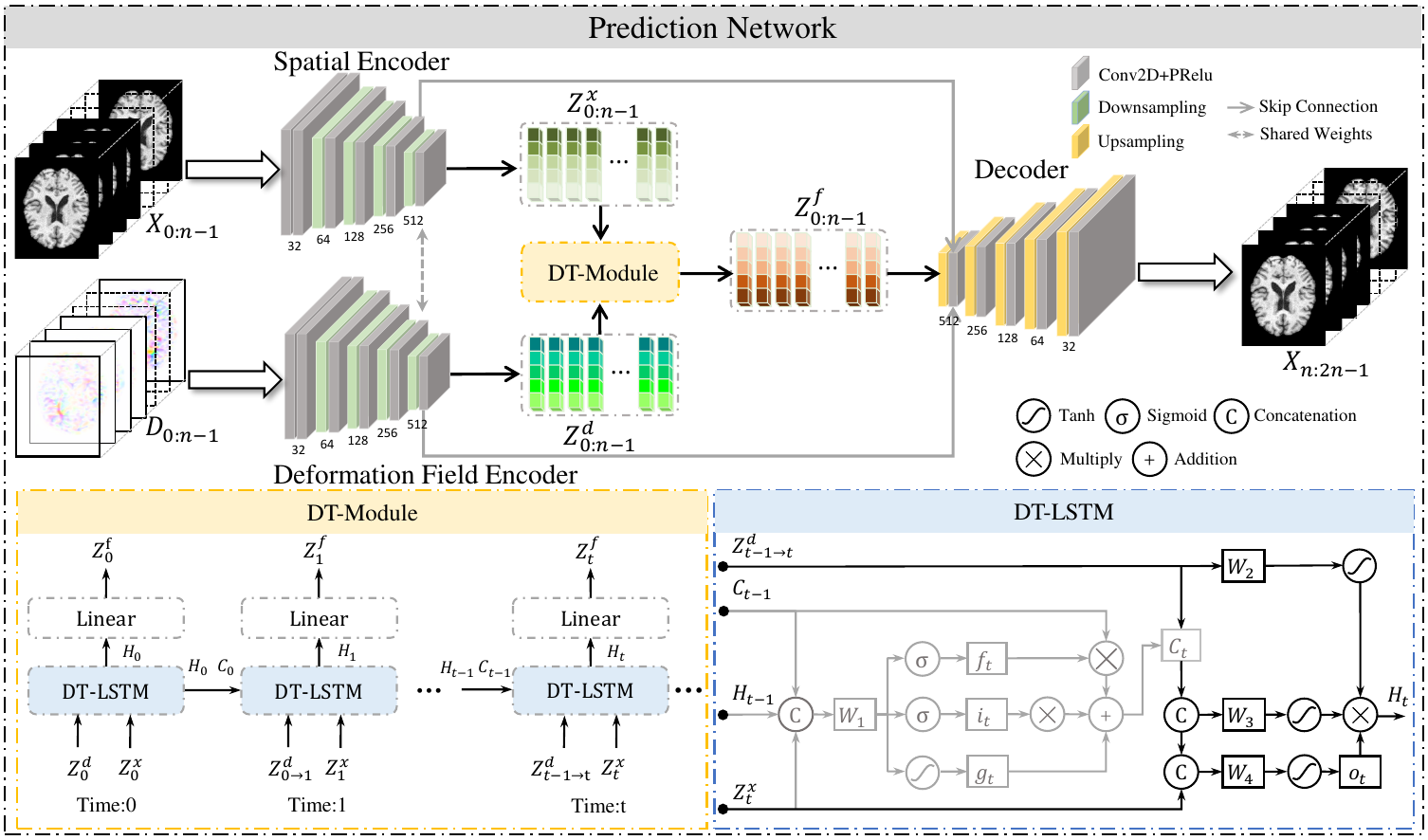}
\caption{The temporal prediction module includes a prediction network, a DT-Module and a DT-LSTM. The gray part represents the prediction network, the yellow part represents DT-Module and the blue part represents DT-LSTM. Among them, the prediction network receives the temporal sequence of brain images $X_{1:n}$ and the temporal sequence of deformation fields $D_{1:n}$, and outputs the temporal sequence of future brain images $X_{n+1:2n}$ after performing temporal modeling. The DT-Module with DT-LSTM are used for temporal modeling. 
}
\label{fig.prediction}
\end{figure*}

The structure of the temporal prediction module is shown in Fig. \ref{fig.prediction}. First, the input sequence $X_{0:n-1}$ is processed using the deformation field estimation network ${E}$ to infer the interframe deformation field sequence $D_{0:n-1}$.  The spatial encoder and the deformation field encoder are then used to extract features from the brain image sequence $X_{0:n-1}$ and the deformation field sequence $D_{0:n-1}$, respectively, and obtain the temporal feature encoding $Z_{0:n-1}^{x}$ of the brain images and the temporal feature encoding $Z_{0:n-1}^{d}$ of the deformation fields. Next, the DT-Module is used to model the temporal correlation between $Z_{0:n-1}^{x}$ and $Z_{0:n-1}^{d}$, generating the fused temporal feature encoding $Z_{0:n-1}^{f}$. Finally, $Z_{0:n-1}^{f}$ is input into the decoder to generate the temporal sequence of future temporal brain images $X_{n:2n-1}$.  

\subsubsection{The Encoder}

The operation of the encoder mainly focuses on the spatial modeling task of the temporal sequence of brain images. In order to help the model better understand the static structural information and the deformation information in the temporal sequence, and to achieve the extraction of deformation features under the constraint of static structural information, a single encoder is used to solve the unimodal features, allowing the network to focus on the feature learning of individual information. 

The spatial encoder focuses on the spatial structure of the brain image itself, learning the spatial distribution of brain image regions, and obtains the brain image feature vector $Z_{0:n-1}^{x}$ after encoding the input temporal sequence of brain images. 

The deformation field encoder focuses on the information of local regions of the deformation fields in the temporal sequence, learning the distribution of deformation regions in the temporal sequence of brain images, and obtains the temporal feature encoding $Z_{0:n-1}^{d}$ of the deformation field after encoding the input deformation field. 

Due to the spatial correlation of static structural information and deformation field information over time, to synchronize their features in terms of spatial modeling capability, the spatial encoder and the deformation field encoder perform feature alignment in space with shared weights between the two. They both adopt the same backbone network structure as the deformation field estimation network E.

\subsubsection{The DT-Module}

The main function of the DT-module is to model the temporal aspects of brain images. The DT-module consists of DT-LSTM and linear layers, which extend the ability to process intraframe deformation fields and temporal relationships from the original LSTM, which can only handle interframe temporal relationships.

The DT-module receives the temporal feature encoding $Z_{0:n-1}^{x}$ from brain images and the temporal feature encoding $Z_{0:n-1}^{d}$ from deformation fields. It uses DT-LSTM to process the temporal feature encoding of brain images and deformation fields. Finally, through linear layers, it achieves deformation perception and temporal feature fusion to obtain the fused output temporal feature encoding $Z_{0:n-1}^{f}$, enabling the temporal prediction module to have deformation-aware capability.

Motivated by the incorporation of a non-linear layer \cite{wang2022predrnn} into recurrent transformations, DT-LSTM evolves from the single input of the original LSTM to two separate inputs at time t: $Z_{t-1\rightarrow t}^{d}$ and $Z_{t}^{x}$, along with the hidden output state $H_{t-1}$ and the cell state $C_{t-1}$ of the previous DT-LSTM module. These components are processed through computational modules to take advantage of the temporal feature encoding of the deformation field $Z_{t-1\rightarrow t}^{d}$, which guides the temporal parameter updates of the brain image feature encoding $Z_{t}^{x}$. This approach aims to prevent temporal memory ${C_{t}}$ from acquiring redundant features and facilitates the propagation of information between successive frames.
Since the hidden output state $H_{t}$ of the LSTM depends on the memory state $C_{t}$ and the output gate $o_{t}$, this means that the memory unit is forced to simultaneously handle both long- and short-term dynamics. Therefore, the modeling capacity of $C_{t}$ limits the overall performance of the model in complex spatio-temporal variations. To enhance the model's perception of complex spatio-temporal variations in brain image disease progression, DT-LSTM performs temporal modeling through the computational module for the interframe deformation field feature encoding, which is generated by the temporal interpolation module. The gray part of the DT-LSTM module in Fig. \ref{fig.prediction} represents the computational module for temporal memory, while the black part represents the computational module for encoding features of the deformation field between frames.

The temporal memory computational module is used to capture and maintain long-term dependencies in sequences at time t: $Z_{t}^{f}$. It is controlled by the temporal memory ${C_{t}}$ through the forget gate ${f_{t}}$, the input gate ${i_{t}}$ and the input modulation gate ${g_{t}}$, where the subscript ${t}$ denotes the time step. The specific parameter update equation is shown in Eq. \ref{eq:dt-lstm1}:   

\begin{equation}
\label{eq:dt-lstm1}
\begin{pmatrix}g_{t}
 \\i_{t}
\\f_{t}
\end{pmatrix} 
=
\begin{pmatrix}\tanh
 \\\sigma
\\\sigma
\end{pmatrix} 
W_{1} * 
[Z_{t}^{x}, H_{t-1}, C_{t-1}]
\end{equation}
where the square brackets denote the tensor concatenation, and the parentheses denote a system of equations. ${g_{t}}$ represents the modulation gate, ${i_{t}}$ represents the input gate, ${f_{t}}$ represents the forget gate, ${W_{1}}$ represents the convolutional kernel parameters, ${*}$ represents the convolution operation, , ${C_{t-1}}$ represents the temporal memory unit of the previous time step, $Z_{t}^{x}$ represents brain image feature encoding, and ${H_{t-1}}$ represents the hidden vector of the previous time step. 

The computational module for inter-frame deformation field feature encoding updates the memory parameters by adding deformation information to the temporal memory. Here, the temporal memory ${C_{t}}$ is updated by the Hadamard product $\odot$ of the forget gate ${f_{t}}$ and the temporal memory of the previous time step ${C_{t-1}}$, plus the Hadamard product $\odot$ of the modulation gate ${g_{t}}$ and the input gate ${i_{t}}$. The equation is shown in Eq. \ref{eq:dt-lstm2}.  

\begin{equation}
\label{eq:dt-lstm2}
C_{t} = f_{t} \odot C_{t-1} + i_{t} \odot g_{t} 
\end{equation}

The output gate ${o_{t}}$ is obtained by convolving ${Z_{t}^{x}}$, the temporal memory ${C_{t}}$ and the feature encoding of the deformation field ${Z_{t-1\rightarrow t}^{d}}$ and ${W_{4}}$, as shown in Eq. \ref{eq:dt-lstm3}.  

\begin{equation}
\label{eq:dt-lstm3}
o_{t} = \tanh (W_{4} * [Z_{t}^{x}, C_{t}, Z_{t-1\rightarrow t}^{d}])
\end{equation}

The hidden vector ${H_{t}}$ is obtained by the operation of the output gate ${o_{t}}$, the deformation feature encoding ${Z_{t-1 \rightarrow t}^{d}}$ and the temporal memory ${C_{t}}$, as shown in Eq. \ref{eq:dt-lstm4}.  

\begin{equation}
\label{eq:dt-lstm4}
H_{t} = o_{t} \odot \tanh(W_{2} * Z_{t-1 \rightarrow t}^{d}) \odot \tanh(W_{3} * [C_{t}, Z_{t-1 \rightarrow t}^{d}])
\end{equation}

As the number of input time steps increases, not only the feature vectors generated by DT-LSTM contain more contextual information at each time step, but also the computational module of deformation field enables DT-LSTM to gain the capability to perceive global temporal morphological changes in a temporal sequence.  
\subsubsection{The Decoder}

To ensure the detailed features of the generated brain images, residual connections are used between the spatial encoder, the deformation encoder, and the decoder to transmit the residuals of spatial and deformation features at each scale in the network to each layer of the decoder to mitigate information loss during the downsampling process of the spatial and deformation encoders.

\subsubsection{The Loss Function}

For the temporal brain image prediction module, the loss $L_{Pred}$ is based on the loss function ${L_2}$, with the aim of obtaining high-quality brain image prediction results by minimizing the loss $L_{Pred}$. The loss of the temporal prediction network $L_{Pred}$ is backpropagated at each time step, and this training method is iteratively performed, as shown in Eq. \ref{eq.pred1}.  

\begin{equation}
\label{eq.pred1}
L_{Pred} = \lVert X_{t}^{gt}-X_{t}^{pred}\rVert_2^2
\end{equation}
where $X_{t}^{gt}$ represents the actual future brain image frame, and $X_{t}^{pred}$ represents the predicted image frame.

\section{The Experiments Results and DISCUSSION}

This paper evaluates the effectiveness and performance of DATGN by collecting longitudinal temporal brain images from the Alzheimer's Disease Neuroimaging Initiative (ADNI) data set \footnote{The dataset can be obtained from https://adni.loni.usc.edu/}.  
This section presents the designed ablation experiments for temporal interpolation and temporal prediction of brain images to verify the effectiveness of deformation information in modeling temporal sequences of brain images.

\subsection{Data Processing}

\begin{figure}[ht]
\centering
\includegraphics[width=\textwidth]{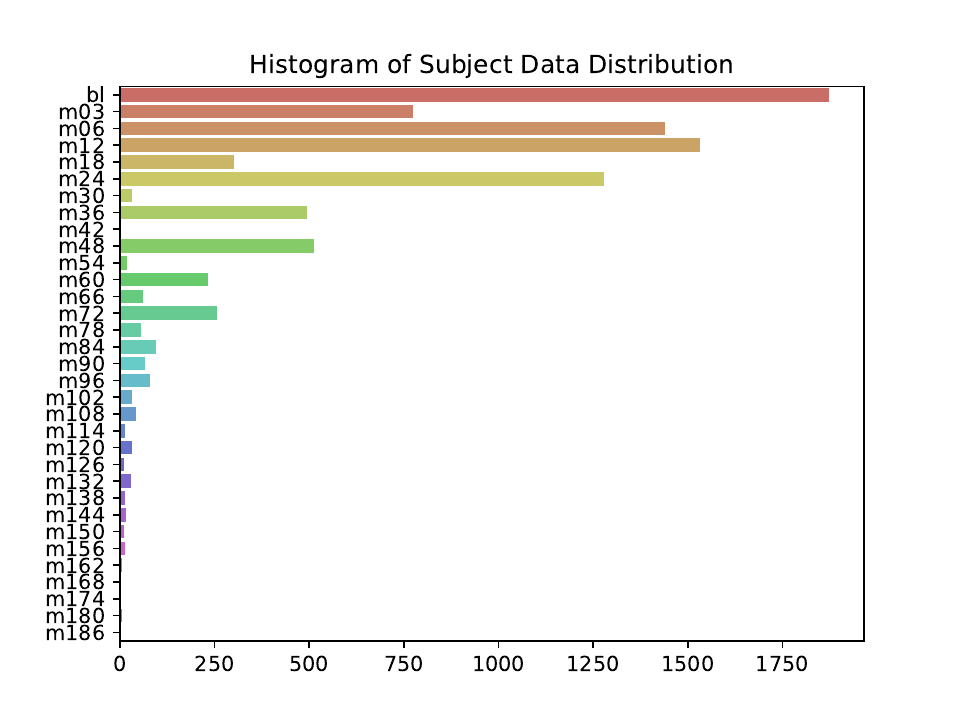}
\caption{A histogram of the distribution of data from 1872 subjects from the ADNI dataset, where the vertical axis represents the time span in months and the horizontal axis represents the sample size.}
\label{data_ana}
\end{figure}

 The ADNI dataset we use is shown in Table \ref{table_1}, which contains a total of 2796 samples. There are 3 classes for all samples, and the maximum subject age is 97 years and the minimum age is 52 years.
 \begin{table}[ht]
\centering
\caption{ADNI dataset distribution}
\label{table_1}


\begin{tabular}{ccccc}

\hline
Origprot & Class      & Samples  \\  \hline
\multirow{3}{*}{ADNI1}    & CN       & 1346  \\
    & AD & 1149  \\
    & MCI      & 1687   \\  \hline
\multirow{3}{*}{ADNI2}    & CN       & 1215  \\
    & AD & 551      \\
    & MCI      & 1336              \\  \hline
\multirow{3}{*}{total}    & CN       & 2561   \\
    & AD & 1700     \\
    & MCI      & 3023                  \\ \hline
\end{tabular}


\end{table}
 The distribution of the data over time is illustrated in Fig. \ref{data_ana}, where ‘m’ on the vertical axis represents the months, 'bl' in the first row indicates the number of subjects who underwent their first examination, and the horizontal axis represents the number of MRI samples. 

This paper selected 1100 samples from 637 subjects, including 2 or more temporal sequences of brain images from 3 to 5 years in ADNI. Consequently, our experiment used 3 years as the threshold to distinguish between short- and long-term data for subjects within 3 years and those between 3 to 5 years, respectively, for separate experimental analysis. 

In terms of data preprocessing, brain images were first segmented using HD-BET\cite{isensee2019automated} and the skull part was removed, followed by head orientation correction and normalization using FSL\cite{woolrich2009bayesian}, and then the brain images were aligned according to the Montreal Neurosciences Institute (MNI152) template. Subsequently, the brain images were proportionally scaled and zero-padded to resize the brain images to $220 \times 220 \times 220$ without deformation. The experiment involved slicing the specimen into 20 units of thickness in three directions, i.e., coronal, sagittal and transverse directions, and then calculating the average.

\subsection{Exprimental Setting}

The methods are trained by 5-fold,  all available samples are divided into the training set and the validation set in a ratio of 8:2. Specifically, 80\% of the data is used for model training and 20\% to evaluate model performance. 
The methods are trained for 200 iterations using the Adam optimizer, with an initial learning rate set to $1 \times 10^{-4}$, and reduced by a factor of 10 every 50 iterations after stabilization. The batch size was set to 8. All models were implemented using the PyTorch framework and trained on a NVIDIA GPU 4090.

\subsection{The Deformation Field Ablation Experiments}

This section demonstrates the effectiveness of deformation information in modeling temporal sequences of brain images through ablation experiment. Firstly, in the temporal interpolation ablation experiment, PSNR and MSE metrics were tested for brain images generated without deformation information, with unidirectional deformation information, and with bidirectional deformation information, demonstrating the impact of deformation information on the quality of brain image generation. Secondly, through the temporal prediction ablation experiment, PSNR and MSE metrics of the baseline model LSTM and the deformation-aware model DT-LSTM were tested to further demonstrate the impact of deformation information on the quality of brain image generation.

\subsubsection{The Deformation Field Ablation Experiments for Brain Image Temporal Interpolation}

To verify whether deformation information can guide the temporal interpolation module to generate more accurate missing brain image frames, this study compared and quantitatively analyzed the results of short- and long-term DATGN brain image generation under no deformation field scene (baseline model), unidirectional deformation field scene and bidirectional deformation scene, as shown in Tab. \ref{tab:ablation_deform}.



In the short-term time span, the experimental results indicate that incorporating deformation field information in short-term temporal interpolation is advantageous to the model to generate more accurate interpolation results. 
Expanding to a long-term time span, the experimental results indicate that incorporating deformation field information also has a significant impact on generating more accurate interpolation results for the model in the long-term time span.  

In summary, DATGN outperforms the baseline model in both metrics on short- and long-term temporal sequences. The experimental results indicate that there is a correlation between the quality of brain images and the deformation characteristics of the temporal sequence of brain images. Clearly, the model can adequately learn the deformation field information of the lesion area over time, enabling the model to take deformation characteristics into account when generating a temporal sequence of future brain images.

\subsubsection{The Ablation Experiments for Brain Image Temporal Prediction}

To demonstrate the impact of the deformation perception ability of the DT module on the quality of brain image generation, Table \ref{tab:ablation_dt} presents the results of the quantitative analysis of DATGN temporal brain image generation without the DT module (LSTM) and with the DT module (DT-LSTM).



From the perspective of data integrity, the experimental results indicate that DT-LSTM can generate more accurate prediction results regardless of data completeness. In the case of missing data in temporal sequences, DT-LSTM infers the missing part of the data and updates the temporal memory parameters through the dynamic deformation-aware module, improving the accuracy of temporal prediction of incomplete brain image sequences. In the case of complete data in temporal sequences, DT-LSTM can better learn the change features within the temporal sequences and improve the accuracy of temporal prediction of complete brain image sequences. Therefore, compared to the traditional LSTM model, DT-LSTM can better capture the temporal features of the data through its dynamic deformation-aware capability, thus achieving more accurate prediction results regardless of the completeness of the data.



From the perspective of the length of the temporal sequence, the experimental results show that DATGN has achieved improvements in the metrics on the task of generating temporal sequences in four dimensions, i.e., short-term sequence, long-term sequence, incomplete sequence, and complete sequence, indicating that DT-LSTM has deformation-aware capability, and deformation awareness is effective for the quality of brain image generation.

\begin{table}[!t]
\caption{Evaluation of Deformation Field in Brain Image Temporal Interpolation 
\label{tab:ablation_deform}}


\begin{tabular}{lcll}
\hline
Method                        & Time span          & PSNR↑   & MSE↓   \\ \hline
Baseline                      & \multirow{3}{*}{Short-term($\leq$ 3 years)}      & 29.3367 & 0.0013 \\
Baseline+Deformation Field    &      & 30.6322 & 0.0011  \\
Baseline+Bi-deformation Field &      & \textbf{30.8694} & \textbf{0.0009}  \\ \hline
Baseline                      & \multicolumn{1}{l}{\multirow{3}{*}{Long-term(\textgreater 3 years)}}      & 24.4358 & 0.0048 \\
Baseline+Deformation Field    &      & 27.7172 & 0.0022 \\
Baseline+Bi-deformation Field &      & \textbf{28.3569} & \textbf{0.0018} \\ \hline
\end{tabular}
\end{table}

\begin{table}[!t]
\caption{Effectiveness of Deformation Field in Brain Image Temporal Prediction 
\label{tab:ablation_dt}}
\begin{tabular}{lccll}
\hline
Method         & Complete data        & Time span & PSNR↑    & SSIM↑ \\ \hline
LSTM           & \ding{55} & \multirow{4}{*}{Short-term($\leq$ 3 years)}   & 28.96191 & 0.865 \\
DT-LSTM & \ding{55} &   & \textbf{31.01446} & \textbf{0.915} \\ 
LSTM           & \ding{51}  &    & 32.16074 & 0.893 \\
DT-LSTM & \ding{51}   &   & \textbf{34.98648} & \textbf{0.927} \\ \hline
LSTM           & \ding{55} & \multirow{4}{*}{Long-term(> 3 years)}   & 25.15694 & 0.766 \\
DT-LSTM & \ding{55} &   & \textbf{29.7908}  & \textbf{0.908} \\ 
LSTM           & \ding{51}   &    & 32.42329 & 0.878 \\
DT-LSTM & \ding{51}   &   & \textbf{33.97529} & \textbf{0.923} \\ \hline
\end{tabular}
\end{table}

\subsection{Evaluation of Short- and Long-term Brain Images Temporal Generation}

To evaluate the quality of DATGN in generating short- and long-term temporal sequences of brain images, this section first evaluates the quality of brain image temporal interpolation based on missing data for short and long-term temporal sequence interpolation tasks. It then compares the quality of generating the temporal sequence of future brain images under data interpolation and data missing conditions in short and long-term temporal sequence prediction tasks. 

In this experiment, the deformation field estimation network and the interpolation network of the temporal interpolation module, as well as the future temporal generation network of the temporal generation module.

\subsubsection{Evaluation of Brain Image Temporal Interpolation }

To test the performance of the DATGN temporal interpolation network to fill uneven missing data as well as long- and short-term sequences, this experiment quantitatively evaluates the quality of the brain images generated by DATGN interpolation by calculating the difference between the brain images generated by DATGN and the true brain images. During the training process, the temporal data of all subjects was first divided into subsequences of three frames each, with no overlap between any frames within a subsequence. The model learns how to estimate the bidirectional deformation field to generate the missing intermediate frame by using the first and last frames in the subsequence. 

The experiments in this section evaluate the quality of the generated temporal coronal, sagittal, and transverse planes of brain images in the event of missing temporal sequences. As shown in the Tab. \ref{tab:comparison-prediction}, the results indicate that the temporal interpolation module outperforms existing temporal generation models in terms of metrics. Since the DATGN model generates intermediate frames by learning bidirectional deformation fields, it can better capture the changes and features within brain imaging data, thereby enhancing the accuracy and image quality of the interpolation. 
The evaluation results show that, the quality assessment on the generated images reflects the excellent performance of DATGN's temporal interpolation network in filling uneven missing data. 

\begin{table*}[]
\caption{\label{tab:comparison-prediction} \textbf{Evaluation of Long- and Short-term Temporal Brain Images Sequential Generation }
}
\begin{adjustbox}{width=14cm,center}

\begin{tabular}{llccc|cccccc}
\hline
\multicolumn{1}{c}{\multirow{3}{*}{Method}} & \multicolumn{1}{c}{\multirow{3}{*}{Time span}} & \multicolumn{3}{c|}{MRI Interpolation}                                                 & \multicolumn{6}{c}{MRI Prediction}                                                                                                                                      \\ \cline{3-11} 
\multicolumn{1}{c}{}                        & \multicolumn{1}{c}{}                           & \multicolumn{3}{c|}{Incomplete longitudinal data}                                  & \multicolumn{3}{c}{Incomplete longitudinal data}                                 & \multicolumn{3}{c}{Complete longitudinal data}                                   \\ \cline{3-11} 
\multicolumn{1}{c}{}                        & \multicolumn{1}{c}{}                           & \multicolumn{1}{l}{MSE↓} & \multicolumn{1}{l}{SSIM↑} & \multicolumn{1}{l|}{PSNR↑} & \multicolumn{1}{l}{MSE↓} & \multicolumn{1}{l}{SSIM↑} & \multicolumn{1}{l}{PSNR↑} & \multicolumn{1}{l}{MSE↓} & \multicolumn{1}{l}{SSIM↑} & \multicolumn{1}{l}{PSNR↑} \\ \hline
Choi et al.\cite{choi2018stargan}                                     & \multirow{5}{*}{Short-term($\leq$ 3 years)}                          & 5.371                     & 0.8336                    & 16.291                    & 6.582                    & 0.7749                    & 16.729                    & 5.973                    & 0.9824                    & 17.508                    \\
Ning et al.\cite{ning2020ldgan}                                       &                          & 2.98                      & 0.8817                    & 18.605                    & 3.916                    & 0.8567                    & 18.929                    & 2.609                    & 0.9912                    & 18.874                    \\
Isola te al.\cite{isola2017image}                                     &                          & 2.18                      & 0.8263                    & 19.183                    & 4.724                    & 0.8861                    & 19.699                    & 3.34                     & 0.9892                    & 19.334                    \\
Kwon et al.\cite{zhu2017unpaired}                                   &                          & 2.582                     & 0.8749                    & 19.729                    & \textbf{3.667}                    & 0.8613                    & 19.866                    & 2.423                    & 0.9915                    & 20.664                    \\
DATGN(Ours)                                  &                          & \textbf{1.916}                     & \textbf{0.9067}                    & \textbf{20.929}                    & 3.763                  & \textbf{0.8886}                    & \textbf{20.156}                  & \textbf{1.931}                    & \textbf{0.9931}                    & \textbf{21.423}                   \\ \hline
Choi et al.\cite{choi2018stargan}                                     & \multicolumn{1}{l}{\multirow{5}{*}{Long-term(\textgreater 3 years)}}                     & 6.12                      & 0.7338                    & 15.261                    & 10.489                   & 0.6356                    & 14.321                    & 8.648                    & 0.7449                    & 16.032                    \\
Ning et al.\cite{ning2020ldgan}                                       &                     & 2.674                     & 0.7899                    & 16.667                    & 5.497                    & 0.7302                    & 16.728                    & 4.328                    & 0.835                     & 16.937                    \\

Isola et al.\cite{isola2017image}                                     &                     & 2.494                     & 0.8209                    & 18.646                    & 8.443                    & 0.6698                    & 18.725                    & 7.668                    & 0.6827                    & 19.495                    \\
Kwon et al.\cite{zhu2017unpaired}                                   &                     & 2.499                     & 0.8356                    & 16.321                    & 5.167                    & 0.7287                    & 17.746                    & 3.289                    & 0.7324                    & 17.122                    \\
DATGN(Ours)                                  &                     & \textbf{2.487}                     & \textbf{0.8502}                    & \textbf{18.728}                    & \textbf{4.201}                    & \textbf{0.8782}                    & \textbf{19.857}                  & \textbf{3.167}                    & \textbf{0.897}                     & \textbf{20.975}                  \\ \hline
\end{tabular}
\end{adjustbox}
\end{table*}

\subsubsection{Evaluation of  Brain Images Temporal Generating}

To achieve the goal of generating long-term sequences from short-term ones in the task of early prediction for Alzheimer's disease, it is necessary to effectively utilize the limited temporal data. Since existing data primarily focus on short-term image sequences, models struggle to capture the disease features of long-term sequences, which poses challenges to the early prediction of Alzheimer's disease. The experiments in this section utilized a frame length in units of 6 frames compared two types of generation results for a time span of 3-5 years in the future, respectively. The experimental results are shown in the Tab. \ref{tab:comparison-prediction}.  
The experimental results indicate that, under a unified time span, DATGN is better at predicting future brain image frames, regardless of the completeness of the image temporal sequences, compared to existing methods. 

\begin{figure*}[ht]
\centering
\includegraphics[width=\textwidth]{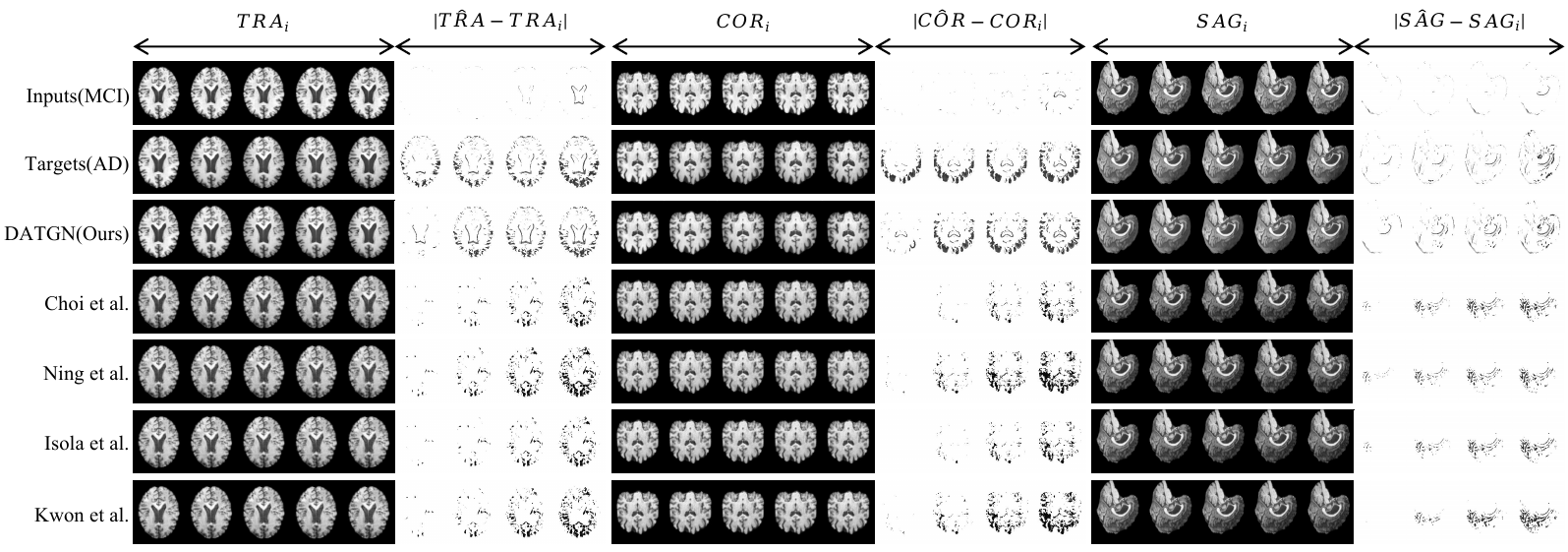}
\caption{Visualization of future MRI and the brain aging process predicted by different methods, with three regions representing MRI slices of the coronal plane (left), sagittal plane (middle) and transverse plane (right), respectively. Within these, each subregion shows the predicted MRI outcomes (left) as compared to the results of the predicted MRI outcomes subtracting the first MRI of the current MRI temporal sequence. }
\label{pred_exp_vis}
\end{figure*}

As brain aging is associated with reduced gray matter volume\cite{avants2008symmetric}, to demonstrate that the deformation field-based temporal modeling is more accurately reflects the true morphological changes associated with aging and disease, the experiments in this section used visualization techniques to present the results of different methods in the temporal sequence generation of brain images, as shown in Fig. \ref{pred_exp_vis}, where $x_{i}$ represents the complete original sequence sampled from the test set, and $|\hat{x}-x_{i}|$ represents each frame subtracting the first frame of the sequence. Each row represents a different method. The experimental results indicate that the images generated by DATGN are more similar to true brain aging in terms of temporal changes, suggesting that DATGN can capture the true morphological changes that characterize the progression of the disease. 


\subsection{Interpolation Ablation of Temporal Image Prediction}

To demonstrate that synthetic data can help the prediction of Alzheimer's disease, this study employed various classifiers to conduct AD/CN and AD/MCI/CN classification experiments on temporal sequences with missing data and complete temporal sequences after adding synthetic data. The accuracy metric was used to measure the discrepancy between the categories predicted by the brain images generated by DATGN and the true disease categories in the five-year prediction to quantitatively evaluate the results of the early prediction. 


\begin{table*}[]
\caption{\label{tab:comparison-prediction-ab} \textbf{Performance evaluation of brain image temporal prediction}
}
\begin{adjustbox}{width=14cm,center}
\begin{tabular}{ccc|cc}
\hline
\multirow{2}{*}{Method} & \multicolumn{2}{c|}{Incomplete Data Accuracy(\%)} & \multicolumn{2}{c}{Complete Data Accuracy(\%)}  \\ \cline{2-5} 
                        & AD/CN Classification  & AD/MCI/CN Classification  & AD/CN Classification & AD/MCI/CN Classification \\ \hline
SVM                     & 41.22\%               & 55.28\%                   & 57.23\%              & 68.24\%                  \\
CNN                     & 56.67\%               & 62.50\%                   & 68.38\%              & 83.75\%                  \\
3D-CNN                  & 66.77\%               & 78.78\%                   & 72.98\%              & 86.12\%                  \\ \hline
\end{tabular}

\end{adjustbox}
\end{table*}

The experimental results show that the accuracy of the DATGN model is significantly improved after the addition of synthetic data to construct complete temporal sequences. This indicates that synthetic data plays an important role in early prediction and provides effective support for the early diagnosis of Alzheimer's disease. This indirectly demonstrates the positive impact of brain imaging data generated by DATGN on early prediction.

\section{Conclusion}
This paper proposes a brain image temporal generation model named DATGN, which utilizes the deformation field information of the images in the course of the disease progression to explore the temporal correlation features based on the deformation information of brain image sequences and generate the future brain images through the deformation fields characterized by changes over time, thereby achieving the early prediction of Alzheimer's disease, assisting doctors and patients in taking interventions and treatment measures and improving the patients' quality of life. 


DATGN consists of two modules, i.e., temporal interpolation and temporal generation. First, we demonstrate the effectiveness of the deformation field in both the temporal interpolation and temporal generation modules for the task of generating temporal sequences of brain images through deformation field ablation experiments. Second, we demonstrated through the quality evaluation experiments of long- and short-term temporal sequence generation of brain images based on deformation field that DATGN can generate brain images that better reflect the brain aging process in the progression of Alzheimer's disease in both short- and long-term sequence generation of brain images in the image interpolation and image prediction experiments. Finally, through data interpolation ablation experiments based on temporal image prediction, we demonstrated that the interpolation algorithm of DATGN can enhance the performance of classification methods. 


In summary, our study presents a novel approach for early prediction of Alzheimer's disease. DATGN can not only generate high-quality interpolated brain images but also predict the patient’s disease progression by generating future brain images of the patient. Future work could further improve our model and algorithm performance and make contributions to early prediction of Alzheimer's disease.



\bibliographystyle{IEEEtran}
\bibliography{references_short}


\vfill

\end{document}